\documentclass[lettersize,journal]{IEEEtran}
\usepackage{amsmath,amsfonts}
\usepackage{array}
\usepackage[caption=false,font=normalsize,labelfont=sf,textfont=sf]{subfig}
\usepackage{textcomp}
\usepackage{stfloats}
\usepackage{url}
\usepackage{verbatim}
\usepackage{graphicx}
\usepackage{cite}
\usepackage[ruled,linesnumbered]{algorithm2e}
\usepackage{booktabs}
\usepackage{makecell,multirow}
\begin{document}

\title{SFSORT: Scene Features-based Simple Online Real-Time Tracker}

\author{M. M. Morsali, Z. Sharifi, F. Fallah, S. Hashembeiki, H. Mohammadzade, S. Bagheri Shouraki}

\markboth{Arxiv preprint, March 2024}%
{Morsali \MakeLowercase{\textit{et al.}}: SFSORT: Scene Features-based Online Real-Time Tracker}

\maketitle

\begin{abstract}
This paper introduces SFSORT, the world's fastest multi-object tracking system based on experiments conducted on MOT Challenge datasets. To achieve an accurate and computationally efficient tracker, this paper employs a tracking-by-detection method, following the online real-time tracking approach established in prior literature. By introducing a novel cost function called the Bounding Box Similarity Index, this work eliminates the Kalman Filter, leading to reduced computational requirements. Additionally, this paper demonstrates the impact of scene features on enhancing object-track association and improving track post-processing. Using a 2.2 GHz Intel Xeon CPU, the proposed method achieves an HOTA of 61.7\% with a processing speed of 2242 Hz on the MOT17 dataset and an HOTA of 60.9\% with a processing speed of 304 Hz on the MOT20 dataset. The tracker's source code, fine-tuned object detection model, and tutorials are available at \url{https://github.com/gitmehrdad/SFSORT}.
\end{abstract}


\section{Introduction}
\IEEEPARstart{M}{ulti-object} tracking involves simultaneously tracking multiple objects in a video, playing a crucial role in applications like autonomous driving, video surveillance, and human-computer interaction. Recently developed high-accuracy object detectors, including Faster R-CNN \cite{fasterrcnn}, Cascaded R-CNN \cite{cascadercnn}, IOU-Net \cite{iounet}, YOLOv3 \cite{yolo}, CenterNet \cite{centernet}, FCOS \cite{fcos}, YOLOX \cite{yolox}, and YOLOv8 \cite{yolo8}, have established tracking-by-detection as the predominant approach in multi-object tracking. In tracking-by-detection, an object detector identifies objects in a frame, and an independent algorithm is employed to associate these detected objects with objects from previous frames. During this association, new IDs are assigned to objects not present in previous frames, while objects existing in prior frames retain their original IDs. The matching between objects detected in a new frame and those from previous frames relies on similarity descriptors such as location cues, motion cues, and appearance cues \cite{sushi}.

The idea behind location similarity is that, in consecutive frames, an object doesn't move much, causing significant overlap in its bounding boxes across frames. Early methods, such as those presented in \cite{ioutracker, sort}, use the Jaccard Index, also known as the intersection over union (IoU), as the cost function for the association. They then use the Hungarian algorithm to find the association with the lowest cost. The problem with IoU is that it gives a zero score for non-overlapping boxes, regardless of their distance; besides, it doesn't account for shape consistency. After the introduction of IoU extensions, like GIoU\cite{giou}, DIoU\cite{diou}, and EIoU\cite{eiou}, which address non-overlapping boxes, some trackers, such as \cite{imprasso, ocsort, strongtbd}, tried using them as their cost function. Meanwhile, other trackers, like \cite{cbioutracker, hybridsort}, proposed different IoU extensions as cost functions for the problem.

While the slight movement assumption usually works well, it might fail in situations with low frame rates, fast movements, or long-term occlusions. Some trackers aim to improve IoU-based association by combining motion and location cues. Assuming linear motion, these trackers used a Kalman Filter (KF) to predict the next location and shape of a bounding box in each frame. Most methods described the bounding box's shape for prediction using the height and aspect ratio \cite{sort, ocsort, strongtbd, deepocsort, deepsort, mat, strongsort, bytetrack}, while others used the height and width \cite{botsort}. The KF employed in some works, including \cite{sort, ocsort, deepocsort}, overlooks the changes in the aspect ratio of a bounding box. Some methods enhance prediction accuracy by combining the detection scores from the object detector with Kalman Filter (KF) results. These approaches employ Noise Scale Adaptive (NSA) Kalman to refine and correct predictions, as demonstrated in \cite{imprasso, strongtbd, strongsort, giaotracker}.
 
The linear motion assumption helps reduce ID switching during short-term occlusions but may not perform well in cases of irregular motion or long-term occlusions. To address this limitation, \cite{ocsort} proposed a method that corrects KF states after reidentifying occluded objects and introduced an angle consistency cost in addition to the standard IoU cost. HybridSORT \cite{hybridsort} further improved OC-SORT \cite{ocsort} by replacing the box center angle consistency cost with an average of bounding box corners cost. Camera movement poses challenges to motion cues, leading to failures of both slight movement and linear motion assumptions. To overcome this challenge, Camera Motion Compensation is necessary. This process involves estimating the camera's rigid motion projection onto the image plane through image registration between consecutive frames. This is achieved by maximizing the Enhanced Correlation Coefficient (ECC) \cite{ecc} or by utilizing features such as sparse optical flow \cite{optflow} or ORB \cite{orb}. Some approaches, as demonstrated by \cite{imprasso, mat, strongsort, invisible, maatrack}, utilize ECC. Others, such as \cite{strongtbd, byte2}, employ ORB features, while some, like \cite{deepocsort, botsort}, rely on sparse optical flow. While ECC offers superior projection accuracy, ORB features and sparse optical flow are often favored for their faster processing speed.

Some trackers use appearance similarity to enhance tracking accuracy \cite{imprasso,strongtbd, hybridsort, deepocsort, strongsort, botsort, smiletrack, poi, lgtrack, motdt}. Typically, these methods employ a reidentification (ReID) deep neural network, such as the ones proposed in \cite{botbaseline, fastreid, wideresnet, dlpar}, to extract appearance embeddings for each bounding box. However, to compare appearances, alternative approaches, like \cite{smiletrack}, use a Siamese network, such as the one introduced in \cite{vit}. Appearance-based methods may yield unreliable results in scenarios involving occluded or blurred detections and similar appearances. Furthermore, the computational load slows them down significantly, making real-time operation impossible. 

A subset of tracking-by-detection methods, such as \cite{sushi, gnmot, mpntrack, gmtracker}, embraces graph-based approaches. These methods utilize graph models to solve the association problem across frames, where graph nodes symbolize detections or feature vectors extracted from them, and graph edges represent connections or similarities between nodes. Despite their potential for high tracking accuracy, graph-based trackers present drawbacks that are incongruent with the objectives of this paper. Firstly, the intricate methodology and the substantial computational load of graph-based trackers often hinder real-time operation. Secondly, many graph-based trackers pursue an offline approach, implying that the association at each frame may depend on subsequent frames.

This paper aims to present a computationally efficient tracker adaptable to various infrastructures, from small edge processor nodes to powerful cloud processor arrays. To design a real-time multi-object tracker balancing accuracy and speed, considering the results of previous works, this paper introduces a motion-aware, location-based multi-object tracking system employing the Hungarian algorithm. The key contributions of this paper are summarized as follows:  

\begin{enumerate}
	\item{This paper presents the Bounding Box Similarity Index (BBSI), a novel similarity descriptor designed to generate association costs for both overlapping and non-overlapping bounding boxes. BBSI takes into account shape similarity, distance, and the overlapping area of bounding boxes.}
	\item{Considering the impact of scene features on both detection and association, the proposed tracker adaptively adjusts its hyperparameters to enhance tracking accuracy.}	
	\item{This paper distinguishes between tracks lost in the video frame's margins and those lost in central areas. Considering different timeouts based on the location of track loss, the presented approach increases the likelihood of revisiting lost tracks.}
	\item{This paper is the first to consider scene features, like scene depth and camera motion, in the post-processing of tracks. It introduces a camera motion detector and an efficient metric for estimating scene depth.}	
\end{enumerate}

\begin{figure*}[!t]
	\centering
	\includegraphics[width=\linewidth]{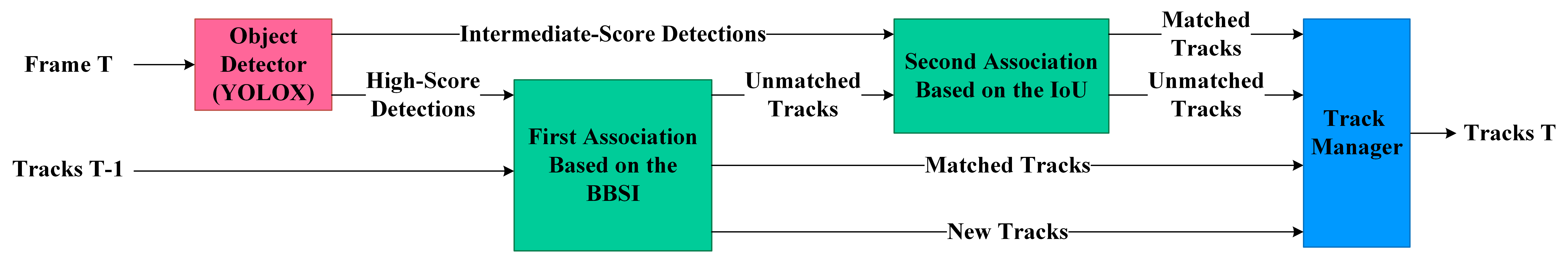}
	\caption{The Proposed Multi-object Tracking System.}
	\label{sfsort}
\end{figure*}

This paper is organized as follows: Section II discusses the proposed method, Section III presents the experiment results, and Section IV concludes the paper.

\section{Proposed Method}
Figure \ref{sfsort} shows the proposed scene features-based online real-time tracker (SFSORT), comprising four components: an object detector, a module for associating high-score detections, a module for associating moderate-score detections, and a track management module. The tracker generates a list of tracks for frame T by processing frame T and the tracks from frame T-1. The following subsections will discuss the functions of each module.

\subsection{Object Detector}
The object detector identifies objects in each video frame, providing their locations and detection scores. The proposed tracking system utilizes the YOLOX object detector, introduced in \cite{yolox}, to attain a remarkable tracking accuracy. The deployed YOLOX model is identical to the one trained and employed in ByteTrack \cite{bytetrack}.

\subsection{First Association Module}
The first association module forms a track pool containing lost and active tracks from frame T-1. Then, based on the Bounding Box Similarity Index, it compares the bounding boxes of tracks in the track pool with the high-score detections from the object detector. Next, it assigns IDs to detections that match with tracks, ensuring that the ID of each object from frame T is the same as its corresponding track from frame T-1. Matched tracks are then sent to the track management module for status updates. High-score detections not matching existing tracks are considered new tracks and forwarded to the track management module for initialization.

\begin{figure*}[!t]
	\centering
	\subfloat[]{\includegraphics[width=1.3in]{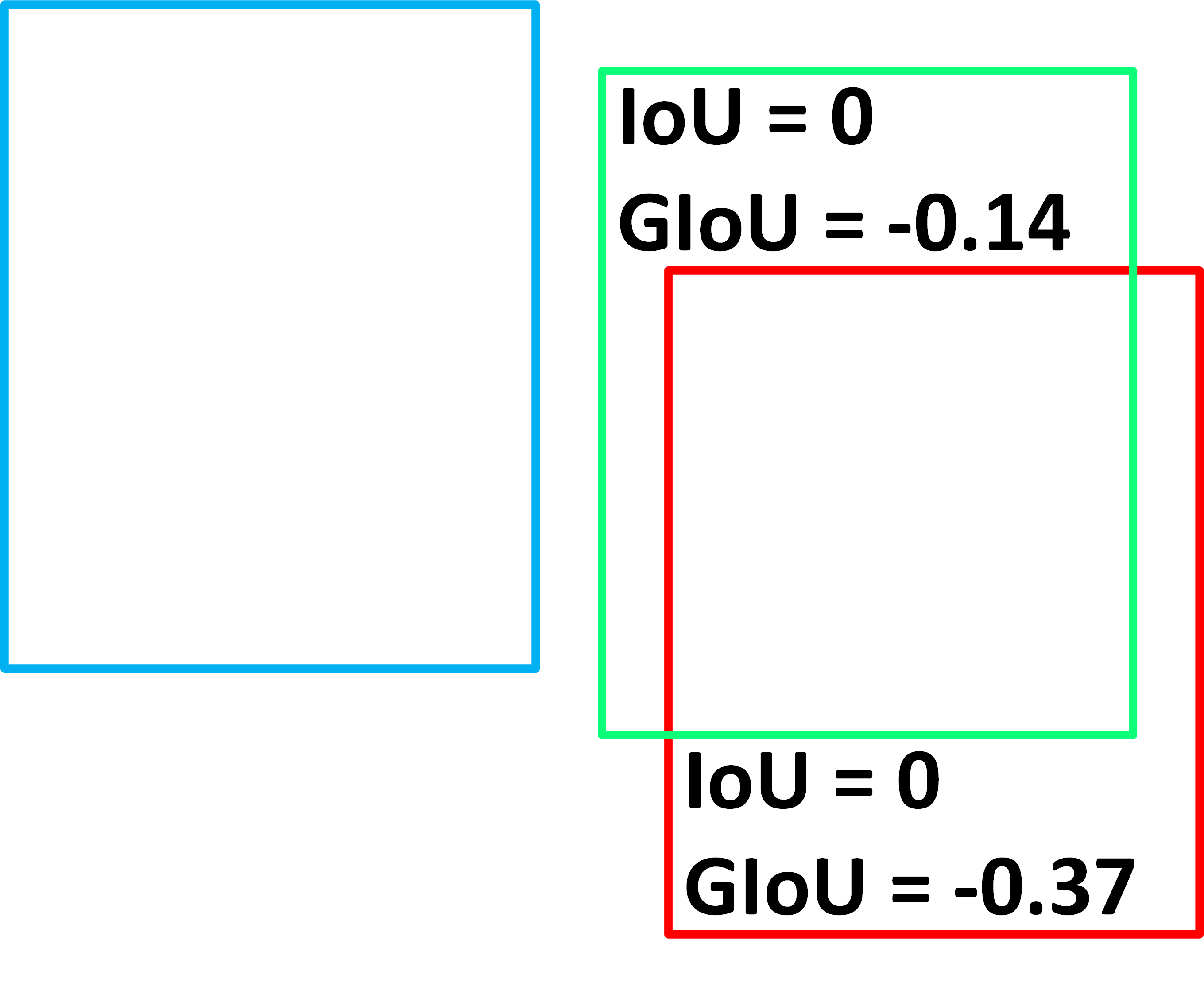}%
		\label{giou}}
	\hfil
	\subfloat[]{\includegraphics[width=0.9in]{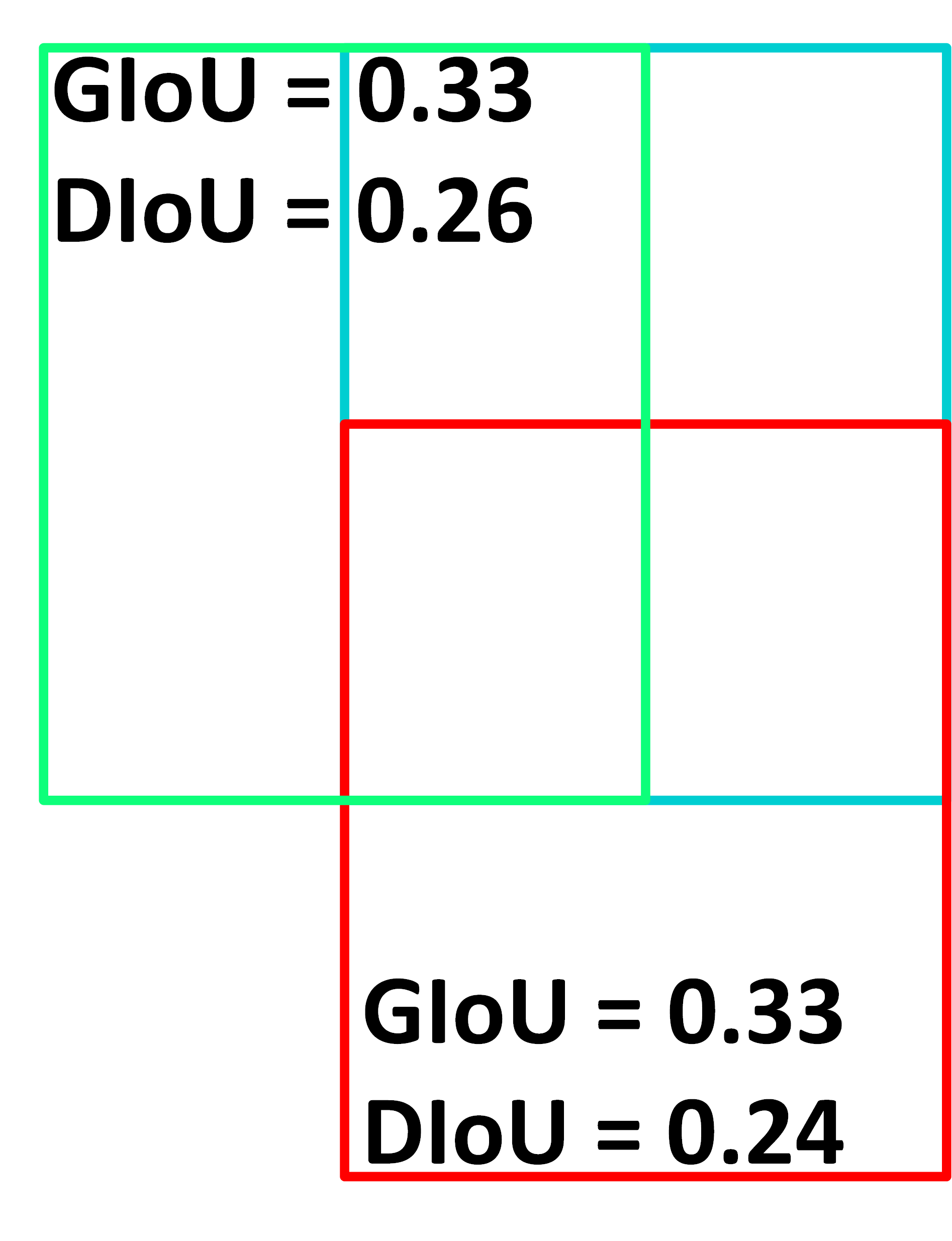}%
		\label{diou}}
	\hfil
	\subfloat[]{\includegraphics[width=0.83in]{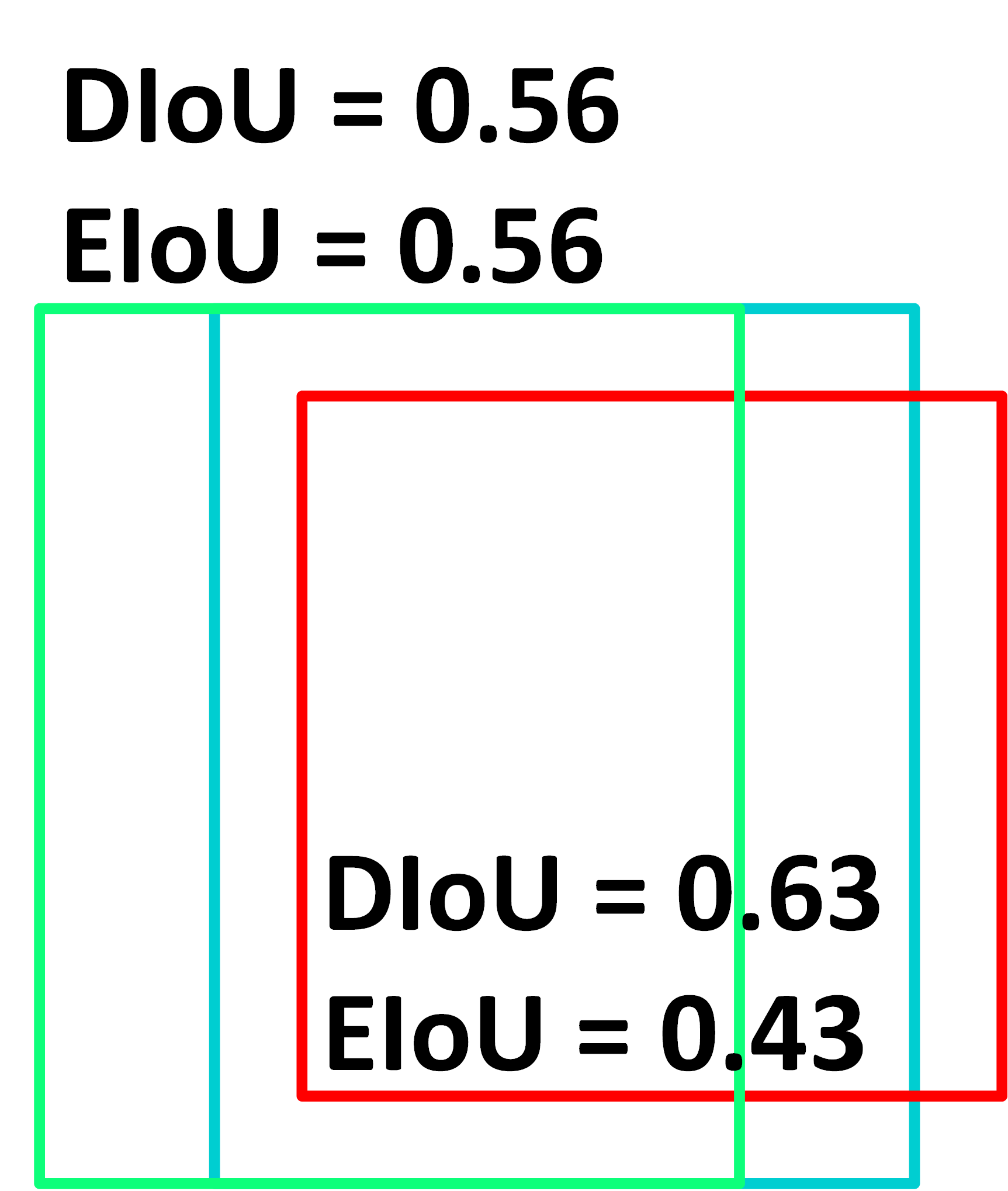}%
		\label{eiou}}
	\caption{Comparison of Various Similarity Descriptors in the Association Problem. (a) IoU vs. GIoU \cite{giou}. (b) GIoU vs. DIoU\cite{diou}. (c) DIoU vs. EIoU\cite{eiou}.}
	\label{descriptors}
\end{figure*}

Figure \ref{descriptors} gives an overview of different similarity descriptors. In this figure, the blue bounding box represents a tracklet, which is the observed bounding box of a track in frame T-1. The green bounding box in Figure \ref{descriptors} indicates an object in frame T that human observation confirms its association with the tracklet from frame T-1. Conversely, the red bounding box in Figure \ref{descriptors} represents an object in frame T that human observation rejects for association with the tracklet from frame T-1. The bounding box for the same object in two consecutive frames may shift position or change size. This can happen due to object detector inaccuracies or the object's motion. Changes in size can occur if the camera moves, the object moves diagonally, or alters its orientation relative to the camera.

The IoU is a popular similarity descriptor, used in \cite{ioutracker, sort, ocsort, bytetrack}, to compare objects and tracks. Equation \ref{eq-iou} defines the IoU:

\begin{equation}
	\label{eq-iou}
	IoU = \frac{A_{Intersection}}{A_{Union}},
\end{equation}
where $A_{Intersection}$ denotes the area of overlap between two bounding boxes, and $A_{Union}$ represents the combined area covered by the two bounding boxes.

As shown in Figure \ref{giou}, IoU yields a score of zero when two bounding boxes don't overlap, regardless of their distance. Consequently, when attempting to associate an object with a track lost a few frames ago, the IoU might fail as the bounding boxes may not overlap due to the object's motion during the loss period. So, some studies suggest using a trajectory prediction tool like a Kalman Filter to consider the motion of bounding boxes \cite{sort, bytetrack, ocsort}. Nevertheless, the errors introduced by trajectory prediction tools and their associated noise may reduce the efficiency of this approach \cite{ocsort}. Therefore, the Generalized IoU (GIoU), introduced in \cite{giou}, is proposed as an alternative similarity descriptor. Equation \ref{eq-giou} shows the GIoU:

\begin{equation}
	\label{eq-giou}
	GIoU = IoU - \frac{A_c - A_{Union}}{A_c},
\end{equation}
where $A_c$ represents the area of the smallest rectangle that encloses both bounding boxes. The term $A_{Union}$, also used in the IoU calculation, represents the total area covered by the two bounding boxes.

As shown in Figure \ref{diou}, GIoU fails to distinguish between two bounding boxes located at different distances from the tracklet. To address this issue, \cite{diou} introduced DIoU, which uses the Euclidean distance between bounding box centers as a similarity measure. Equation \ref{eq-diou} shows the DIoU:

\begin{equation}
	\label{eq-diou}
	DIoU = IoU - \frac{d(box_1, box_2)^2}{h_c^2 + w_c^2},
\end{equation}
where the term $d(box_1, box_2)$ represents the Euclidean distance between the centers of the bounding boxes, while $h_c$ and $w_c$ respectively denote the height and width of the smallest rectangle that encloses both bounding boxes.

In \cite{diou}, the authors also introduced CIoU, an extension of DIoU that accounts for the similarity of bounding boxes' aspect ratios. However, the aspect ratio does not help distinguish objects in most multi-object tracking scenarios, as numerous objects have similar aspect ratios.

As shown in Figure \ref{eiou}, DIoU does not consider variations in bounding box dimensions when evaluating similarity. To overcome this limitation, \cite{eiou} introduced EIoU, which includes the consistency of bounding boxes' width and height. Equation \ref{eq-eiou} demonstrates the EIoU:

\begin{equation}
	\label{eq-eiou}
	EIoU = DIoU - \frac{\Delta h(box_1, box_2)^2}{h_c^2} - \frac{\Delta w(box_1, box_2)^2}{w_c^2},
\end{equation}
where $\Delta h(box_1, box_2)$ shows the height difference between the bounding boxes, and $\Delta w(box_1, box_2)$ represents their width difference. Also, $h_c$ and $w_c$ represent the height and width of the smallest rectangle that encloses both bounding boxes.

\begin{figure}[!t]
	\centering
	\subfloat[]{\includegraphics[width=1.8in]{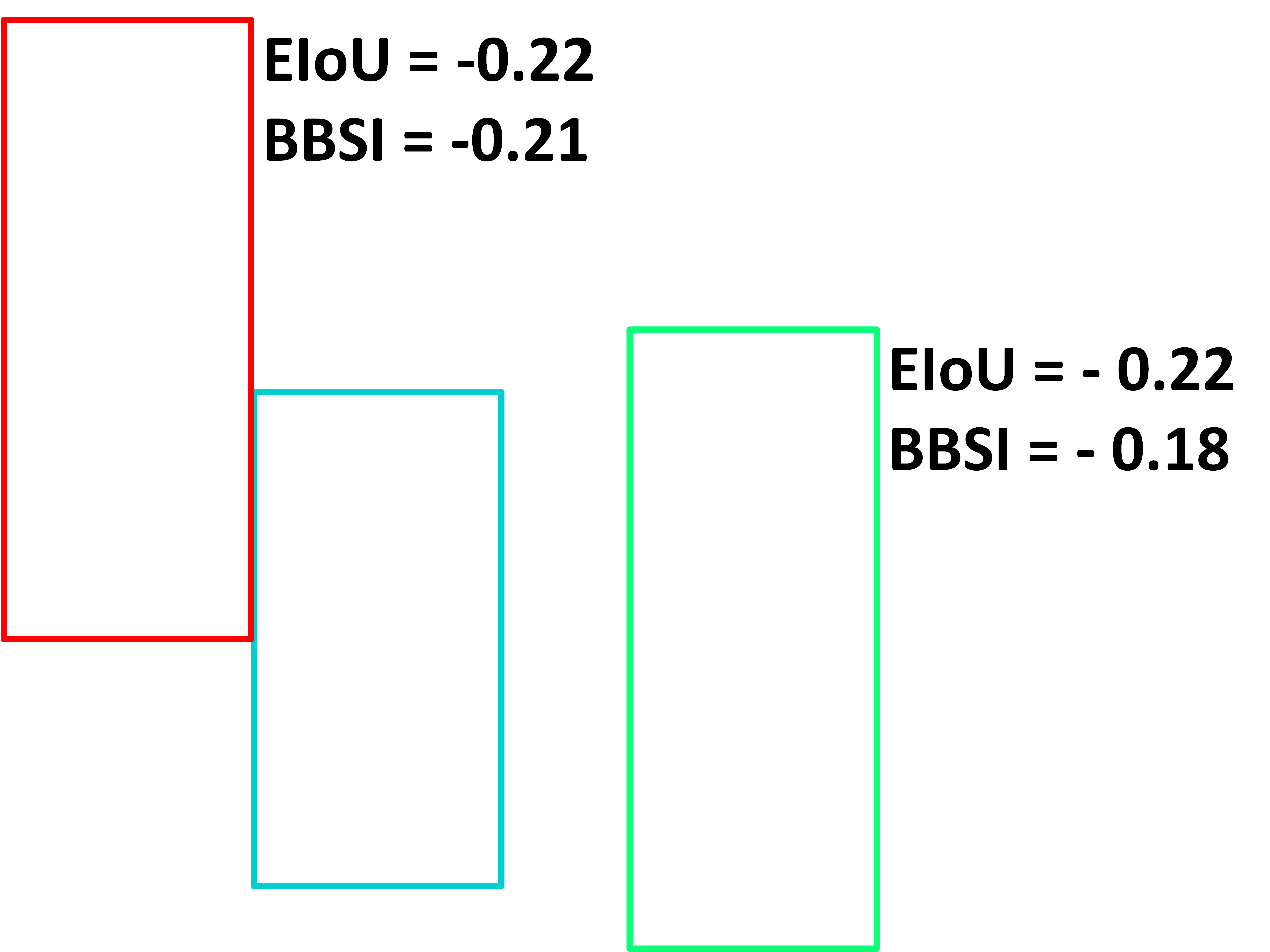}%
		\label{eiouvsbbsi}}
	\vfil
	\subfloat[]{\includegraphics[width=1.8in]{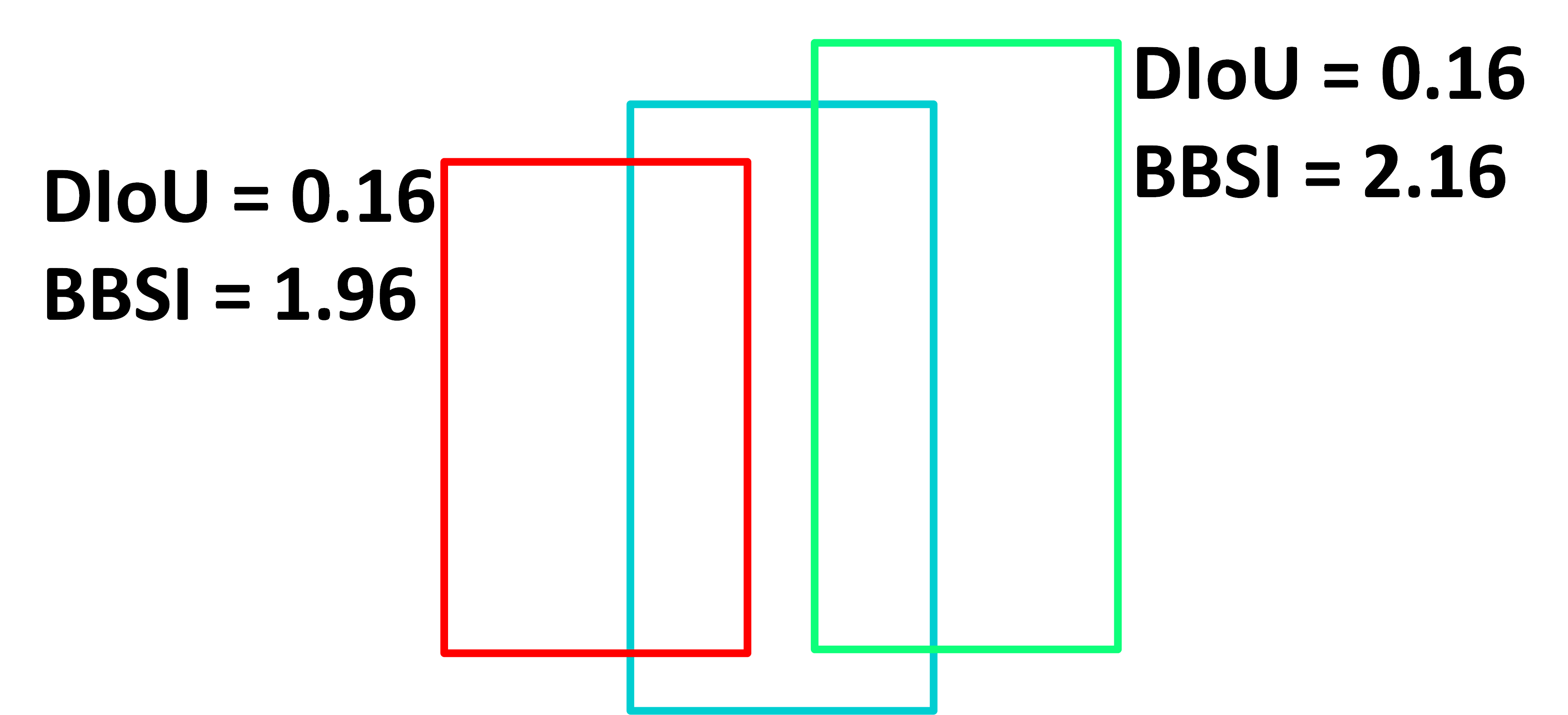}%
		\label{diouvsbbsi}}
	\caption{Comparison of the BBSI with Other Similarity Descriptors. (a) BBSI vs. EIoU\cite{eiou}. (b) BBSI vs. DIoU\cite{diou}.}
	\label{ourmetric}
\end{figure}

As shown in Figure \ref{eiouvsbbsi}, EIoU fails to appropriately associate the tracklet with the bounding box having minimal center-to-center distance due to its sensitivity to the motion direction of the bounding box. This sensitivity stems from the normalization of height and width differences using the height and width of the smallest rectangle that encloses both bounding boxes, respectively. The Bounding Box Similarity Index (BBSI), defined by Equation system \ref{eq-BBSI}, effectively overcomes this limitation.
\begin{subequations}\label{eq-BBSI}
	\begin{align}
		x_{bottom} &= \min(x_{bbox1, rb}, x_{bbox2, rb}). \label{eq-BBSI-A}\\
		x_{top} &= \max(x_{bbox1, tl}, x_{bbox2, tl}). \label{eq-BBSI-B}\\
		y_{bottom} &= \min(y_{bbox1, rb}, y_{bbox2, rb}). \label{eq-BBSI-C}\\
		y_{top} &= \max(y_{bbox1, tl}, y_{bbox2, tl}). \label{eq-BBSI-D}\\
		h_{eff} &= \max(0, x_{bottom} -  x_{top}). \label{eq-BBSI-E}\\
		w_{eff} &= \max(0, y_{bottom} -  y_{top}). \label{eq-BBSI-F}\\
		S_h &= \frac{h_{eff}}{h_{eff} + |h_{bbox2}-h_{bbox1}| +\epsilon}. \label{eq-BBSI-G}\\
		S_w &= \frac{w_{eff}}{w_{eff} + |w_{bbox2}-w_{bbox1}| +\epsilon}. \label{eq-BBSI-H}\\
		S_c &= \frac{|x_{bbox1, c} - x_{bbox2, c}| + |y_{bbox1, c} - y_{bbox2, c}|}{h_c + w_c}. \label{eq-BBSI-I}\\
		ADIoU &= IoU - S_c.\label{eq-BBSI-J}\\
		BBSI &= ADIoU + S_h + S_w. \label{eq-BBSI-K}	
	\end{align}
\end{subequations}

\begin{figure}[!t]
	\centering
	\includegraphics[width=2.7in]{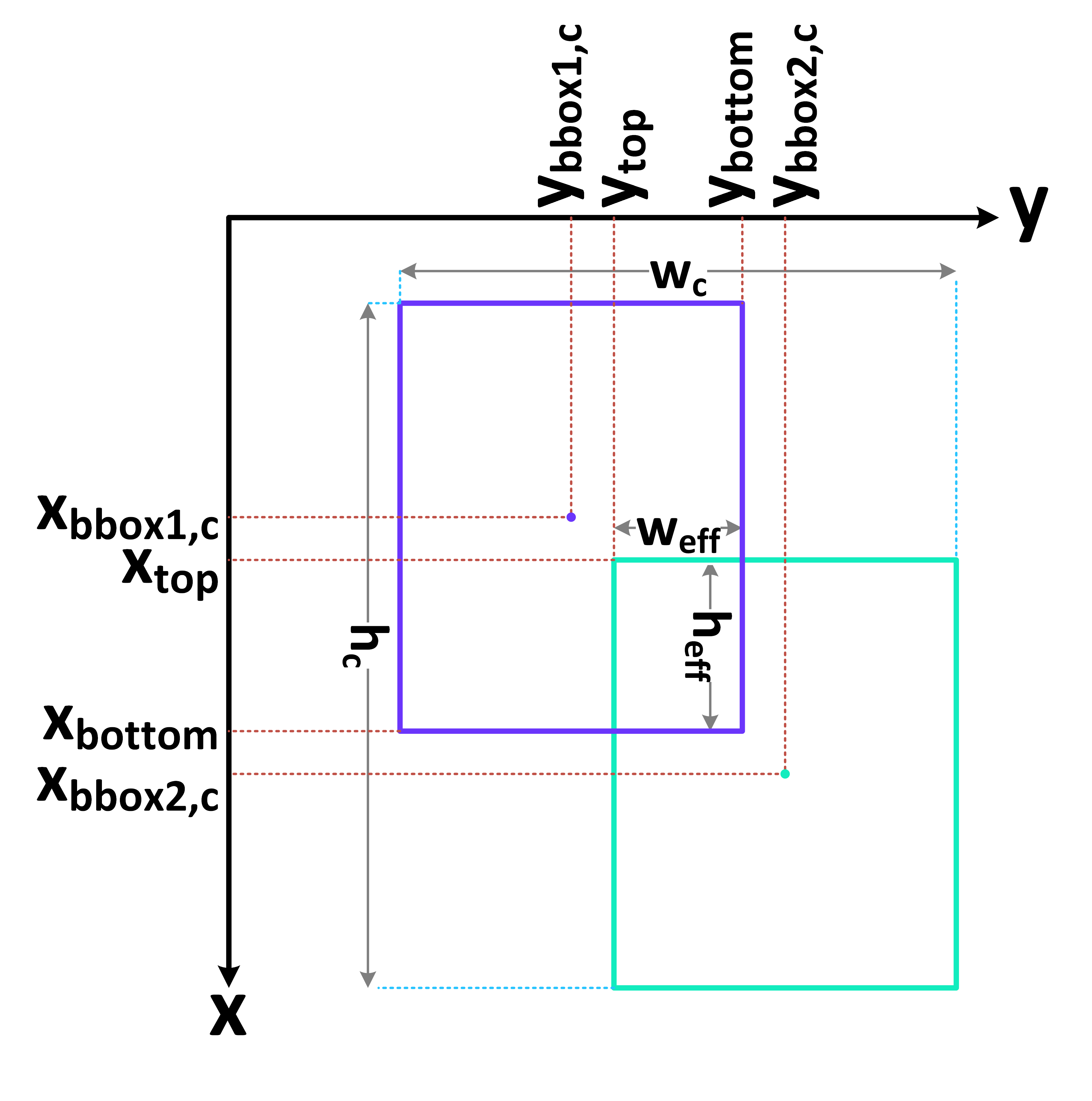}
	\caption{The Visualization of Calculation Details in the BBSI.}
	\label{fig_bbsi}
\end{figure}

Figure \ref{fig_bbsi} visualizes some of the calculation details used in Equation system \ref{eq-BBSI}. In Equation \ref{eq-BBSI-A}, $x_{bbox1, rb}$ denotes the x-coordinate of the right-bottom corner of the first bounding box, and in Equation \ref{eq-BBSI-B}, $x_{bbox1, tl}$ represents the x-coordinate of the top-left corner of the first bounding box. Similarly, the notations used in Equation \ref{eq-BBSI-C} and Equation \ref{eq-BBSI-D} describe the y-coordinates of bounding boxes. In Equation \ref{eq-BBSI-E}, $h_{eff}$ corresponds to the height of the intersection area of two bounding boxes if they overlap; otherwise, it is equal to zero. Similarly, in Equation \ref{eq-BBSI-F}, $w_{eff}$ corresponds to the width of the intersection area. Equation \ref{eq-BBSI-G} uses $h_{eff}$ to define $S_h$, a measure of height similarity, which approaches one for similar bounding boxes and zero for dissimilar ones. Similarly, Equation \ref{eq-BBSI-H} uses $w_{eff}$ for width similarity, denoted as $S_w$. The parameter $\epsilon$ in both Equation \ref{eq-BBSI-G} and Equation \ref{eq-BBSI-H} prevents division by zero. In Equation \ref{eq-BBSI-I}, $x_{bbox1, c}$ represents the x-coordinate of the center of the first bounding box, and $S_c$ denotes the conformity of bounding box centers. Although $S_c$ closely resembles the concept in DIoU, it has been revised to reduce computational overhead. So, Equation \ref{eq-BBSI-J} introduces the ADIoU as an approximation of DIoU. Finally, Equation \ref{eq-BBSI} defines the Bounding Box Similarity Index (BBSI). Figure \ref{diouvsbbsi} illustrates how the BBSI contributes to achieving the correct object-track association when DIoU is insufficient.

The BBSI's consideration of non-overlapping bounding boxes in the association problem diminishes the necessity for motion prediction. Consequently, motion prediction tools such as the Kalman Filter can be eliminated from the tracking system to save computational resources and improve tracking speed.

Consistent with prior research, this paper treats the association problem as a cost-minimization challenge. It begins by constructing a cost matrix, where each element $cost_{ij}$ indicates the cost of associating detection $i$ with track $j$. The cost function employed by the first association module is defined in Equation \ref{eq-cost1}. When two bounding boxes are most similar, they share the same dimensions, resulting in $S_{h}$ and $S_{w}$ both being equal to $1$ in BBSI. Additionally, the coincident centers of these bounding boxes lead to ADIoU being $1$ in BBSI. Consequently, the maximum achievable BBSI is $3$. Conversely, for two entirely dissimilar bounding boxes, $S_{h}$ and $S_{w}$ become $0$, and ADIoU approaches $-1$. Hence, the minimum possible BBSI is $-1$. Therefore, there is a need for a normalization operation to confine the cost to the $(0, 1)$ range. Equation \ref{eq-cost1} demonstrates a computationally-efficient normalization achieved by just one division by 3 operation. This normalization precisely limits the cost to the $(0, 0.66)$ range. After forming the cost matrix, the Jonker-Volgenant algorithm \cite{lapjv}, a high-performance implementation of the Hungarian algorithm, solves the cost minimization problem.

\begin{equation}
	\label{eq-cost1}
	cost_{1st. association} = 1 - \frac{BBSI}{3}.
\end{equation}

\subsection{Second Association Module}
The second association module in Figure \ref{sfsort} is responsible for associating unmatched tracks from the first association with intermediate-score detections. The reduction in detection score often occurs due to partial occlusion of some object parts \cite{bytetrack}, leading to changes in the detection's bounding box dimensions. Consequently, neglecting the similarity of bounding box dimensions, the second association module relies solely on the IoU. Successful matches in the second association module are then forwarded to the Track Manager for status updates.

The cost function employed by the second association module is represented in Equation \ref{eq-cost2}, limiting the cost to the $(0, 1)$ range.

\begin{equation}
	\label{eq-cost2}
	cost_{2nd. association} = 1 - IoU.
\end{equation}

\subsection{Track Manager}
After the association process, the Track Manager, depicted in Figure \ref{sfsort}, handles matched tracks, new tracks, and unmatched tracks. For matched tracks, the Track Manager updates the track's status to \emph{active}, records its bounding box coordinates, and registers the number of the frame containing the track. For unmatched tracks, the Track Manager examines the track's last recorded bounding box coordinates, and depending on where the track was lost, it modifies the track's status to either \emph{lost-at-center} or \emph{lost-at-margin}. The tracking system retains lost tracks in anticipation of their potential revisit. Nevertheless, the Track Manager removes tracks that have been lost for a duration exceeding a specified time-out from the list of tracks eligible for the object-track association. Notably, the time-out for \emph{lost-at-margin} tracks is shorter than that for \emph{lost-at-center} tracks. This distinction arises from the understanding that when a track is lost at the periphery of a frame, it is more likely due to the object moving out of the camera's field of view. In contrast, when a track is lost at the central portion of a frame, it is often attributed to occlusion or blurring. Hence, this paper proposes separate revisiting time-outs for marginal and central tracks.

\subsection{The Proposed Algorithm}
Algorithm \ref{alg:sfsort} shows the proposed multi-object tracking method. It takes object bounding boxes and their detection scores from the object detector and produces the resulting video tracks. After processing each video frame, the algorithm updates two lists: \emph{Active Tracks} and \emph{Lost Tracks}. Between Line 2 and Line 9, the algorithm removes any lost track that has exceeded its allowed time-out from the \emph{Lost Tracks}. In Line 10, the \emph{Lost Tracks} and \emph{Active Tracks} combine to form the \emph{Track Pool}. Between Line 11 and Line 18, the algorithm divides detections into two categories of \emph{Definite Detections} and \emph{Possible Detections} based on their detection scores. The first object-track association occurs between Line 19 and Line 25, involving \emph{Definite Detections} and tracks from the \emph{Track Pool}. During this first association, at Line 21, the algorithm updates the status of matched tracks to the active state, and it records their bounding box along with the current frame number. Additionally, at Line 22, the algorithm removes any matched tracks present in the \emph{Lost Tracks} list. Unmatched high-score detections are identified as new tracks and undergo an initialization process similar to the activation process of matched tracks, as detailed in Line 23. The algorithm adds all associated tracks during the first association to the \emph{Active Tracks} list at Line 24. Between Line 26 and Line 29, when the \emph{Track Pool} is empty, as is often the case at the beginning of object tracking, the algorithm identifies all high-scoring detections as new tracks, initializes them, and adds them to the \emph{Active Tracks} list.

\begin{algorithm}[!t]
	\caption{Pseudo-code of SFSORT.}\label{alg:sfsort}
	\KwIn{$Box$: bounding boxes of objects\; $Score$: detection scores\; $HTH$: high-score detection score threshold\; $LTH$: intermediate-score detection score threshold\; $CTime$: central lost track time-out\; $MTime$: marginal lost track time-out} 
	\KwOut{$Tracks$: video tracks}
	\For{$FrameNo\gets1$ \KwTo $T$ }
	{
		\For{$Track$ in LostTracks}
		{
			\If{$Track$.Status=Central and ($FrameNo$ - $Track$.LastFrameNo $>CTime$)}
			{
				LostTracks.Remove($Track$);
			}
			\If{$Track$.Status=Marginal and ($FrameNo$ - $Track$.LastFrameNo $>MTime$)}
			{
				LostTracks.Remove($Track$);
			}				
		}
		TrackPool $\gets$ ActiveTracks + LostTracks\;
		\For{$DetectionNo\gets1$ \KwTo $N$ }
		{
			\If{$Score[DetectionNo] > HTH$}
			{
				DefiniteDetections.Append($Box[DetectionNo]$);
			}
			\ElseIf{$Score[DetectionNo] > LTH$}
			{
				PossibleDetections.Append($Box[DetectionNo]$);
			}		
		}
		\If{TrackPool is not empty}
		{			
			Matches, UnmatchedTracks, UnmatchedDetections = Associate(TrackPool, DefiniteDetections)\;
			Activate(Matches)\;
			LostTracks.Remove(Matches)\;
			NewTracks = Initialize(UnmatchedDetections)\;
			ActiveTracks.Append(Matches, NewTracks);
		}
		\Else
		{
			 NewTracks = Initialize(DefiniteDetections)\;
			 ActiveTracks.Append(NewTracks);
		}
		\If{UnmatchedTracks is not empty}
		{
			Matches, UnmatchedTracks, UnmatchedDetections = Associate(UnmatchedTracks, PossibleDetections)\;
			Activate(Matches)\;
			ActiveTracks.Append(Matches)\;
			LostTracks.Remove(Matches)\;
			LostTracks.Append(UnmatchedTracks)\;	
			UpdateStatus(LostTracks);	
		}	
		$Tracks$.Append(ActiveTracks);
	}
	\Return{$Tracks$}
	
\end{algorithm}

From Line 30 to Line 37, the second association matches \emph{Possible Detections} and unmatched tracks from the first association. In Line 32 to Line 33, the tracks matched during the second association are activated and added to the \emph{Active Tracks} list. Matched tracks, if present in the \emph{Lost Tracks}, get removed from \emph{Lost Tracks} at Line 34. After the second association attempt, the algorithm adds tracks that remain unmatched to the \emph{Lost Tracks} list in Line 35. In line 36, for lost tracks, the algorithm updates their status, records their last observed bounding box, and registers the track's most recent frame number. The status is determined based on where the track was lost, resulting in either \emph{lost-at-center} or \emph{lost-at-margin}.

At Line 38 of Algorithm \ref{alg:sfsort}, the \emph{Active Tracks} are appended to the \emph{Tracks} list, a collection of tracks from all video frames. Finally, the algorithm returns the \emph{Tracks}.

\subsection{Tuning Hyperparameters Based on Scene Features}
\label{features}
To complete the implementation of a multi-object tracking system, a critical task is to propose a strategy for adjusting hyperparameters. SFSORT comprises nine major hyperparameters indicated in Table \ref{tab_hypers}. The values for these hyperparameters would be determined based on the evaluation results from the tracker's validation set.

\begin{table}
	\begin{center}
		\caption{Major Hyperparameters of SFSORT}	
		\label{tab_hypers}
		\begin{tabular}{cc}
			\toprule
			Symbol & Description \\
			\midrule
			HTH & Minimum score for high-score detections \\
			LTH & Minimum score for intermediate-score detections \\
			MTH1 & Maximum allowable cost in the first association module \\
			MTH2 & Maximum allowable cost in the second association module \\
			NTH & Minimum score for detections identified as new tracks \\
			HMargin & Margin to determine the horizontal boundaries of central areas\\
			VMargin & Margin to determine the vertical boundaries of central areas\\
			CTime & Time-out for tracks lost at central areas\\
			MTime & Time-out for tracks lost at marginal areas\\
			\bottomrule
		\end{tabular}
	\end{center}
\end{table}

In Table \ref{tab_hypers}, \emph{LTH} and \emph{MTH2} are hyperparameters influenced by the behavior of the object detector. Since different object detectors may score intermediate detections differently, it is necessary to adjust these values for each detector selected as the system's object detector. Once adjusted, these values remain constant throughout the tracking process.

The metadata of the input video impacts \emph{HMargin}, \emph{VMargin}, \emph{CTime}, and \emph{MTime} from Table \ref{tab_hypers}. As depicted in Equations \ref{eq-ctime} and \ref{eq-mtime}, this study assumes a linear relationship between the frame rate of the input video and \emph{CTime} and \emph{MTime}, respectively.

\begin{equation}
	\label{eq-ctime}
	CTime = Timeout_{central} \times FrameRate.
\end{equation}

\begin{equation}
	\label{eq-mtime}
	MTime = Timeout_{marginal} \times FrameRate.
\end{equation}

Equation \ref{eq-hm} indicates that the width of the input video's frame linearly affects \emph{HMargin}, and Equation \ref{eq-vm} suggests that the height of the input video's frame has a linear impact on \emph{VMargin}.

\begin{equation}
	\label{eq-hm}
	HMargin = Margin_{horizontal} \times Width.
\end{equation}

\begin{equation}
	\label{eq-vm}
	VMargin = Margin_{vertical} \times Height.
\end{equation}

The count of objects in each video frame influences \emph{HTH}, \emph{NTH}, and \emph{MTH1}. As scenes become more crowded, detection confidence scores typically decrease due to increased occlusion. Consequently, both \emph{HTH} and \emph{MTH1} should be decreased to achieve an adaptive tracker. Moreover, as scenes become more crowded, the number of lost tracks typically increases due to the occlusion. Therefore, \emph{NTH} should be increased to prevent identity switching. Based on two recent conclusions, this paper suggests a linear relationship between the logarithm of the object count in a frame and the hyperparameters \emph{HTH}, \emph{NTH}, and \emph{MTH1}, respectively shown in Equations \ref{eq-hth}, \ref{eq-nth} and \ref{eq-mth}.

\begin{subequations}\label{eq-dyn}
	\begin{align}
		count &= \log_{10} (|\{x| score(x) > CTH\}|). \label{eq-count}\\
		HTH &= HTH_{0} - (HTH_{m} \times count). \label{eq-hth}\\
		NTH &= NTH_{0} + (NTH_{m} \times count). \label{eq-nth}\\
		MTH1 &= MTH_{0} - (MTH_{m} \times count). \label{eq-mth}
	\end{align}
\end{subequations}

In Equation \ref{eq-count}, the count of objects in the scene is determined by the number of set members whose detection score exceeds a certain threshold. This introduces the threshold, denoted as CTH, as a minor hyperparameter in the tracking problem. Equation \ref{eq-hth} introduces two additional minor hyperparameters into the problem, denoted as $HTH_{0}$ and $HTH_{m}$, representing the intercept and slope of a linear equation respectively. Similarly, Equations \ref{eq-nth} and \ref{eq-mth} each introduce two minor hyperparameters to the tracking problem. The values of $HTH_{0}$ and $NTH_{0}$ are restricted within the range $(0, 1)$, while $MTH_{0}$ is confined within the range $(0, 0.66)$.

\subsection{Post-processing}
\label{post}
In cases where online tracking is not necessary, post-processing can enhance tracking accuracy. A key step in this enhancement includes removing short tracks, typically arising from false detections. Since this paper emphasizes video features as a significant factor in tracking systems, it assumes a linear relationship between the video's frame rate and the minimum length of a valid track. Equation \ref{eq-nmin} illustrates this relationship:

\begin{equation}
	\label{eq-nmin}
	N_{min} = C_{m} \times FrameRate,
\end{equation}

where $N_{min}$ represents the minimum length of a valid track, while $C_{m}$ denotes the coefficient to be determined through experiments conducted on the validation set.

Another post-processing embraces recovering the missed bounding boxes of a revisited lost track using linear interpolation. In this work, interpolation occurs only for tracks lost within a time-out period, as interpolation may predict an erroneous trajectory for tracks lost for a long time. As illustrated in Equation \ref{eq-ndti}, the time-out period is assumed to have a linear relationship with the video's frame rate:

\begin{equation}
	\label{eq-ndti}
	N_{dti} = C_{d} \times FrameRate,
\end{equation}

where \ref{eq-ndti}, $N_{dti}$ represents the time-out period, while $C_{d}$ denotes the coefficient to be determined through experiments conducted on the validation set.

\begin{figure*}[!t]
	\centering
	\subfloat[]{\includegraphics[width=3.5in]{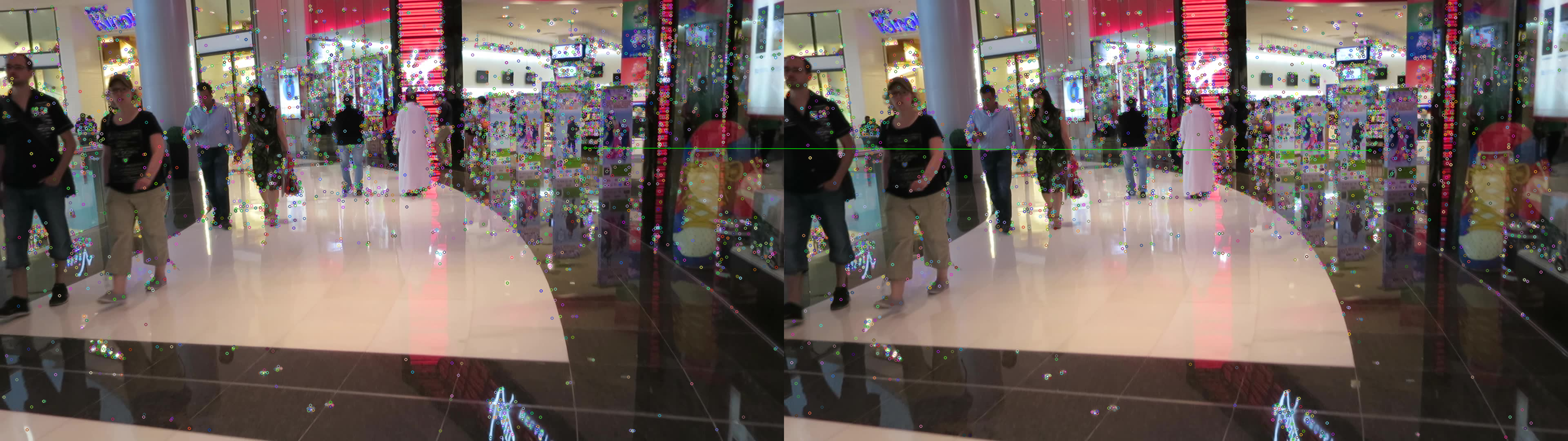}%
		\label{moving}}
	\hfil
	\subfloat[]{\includegraphics[width=3.5in]{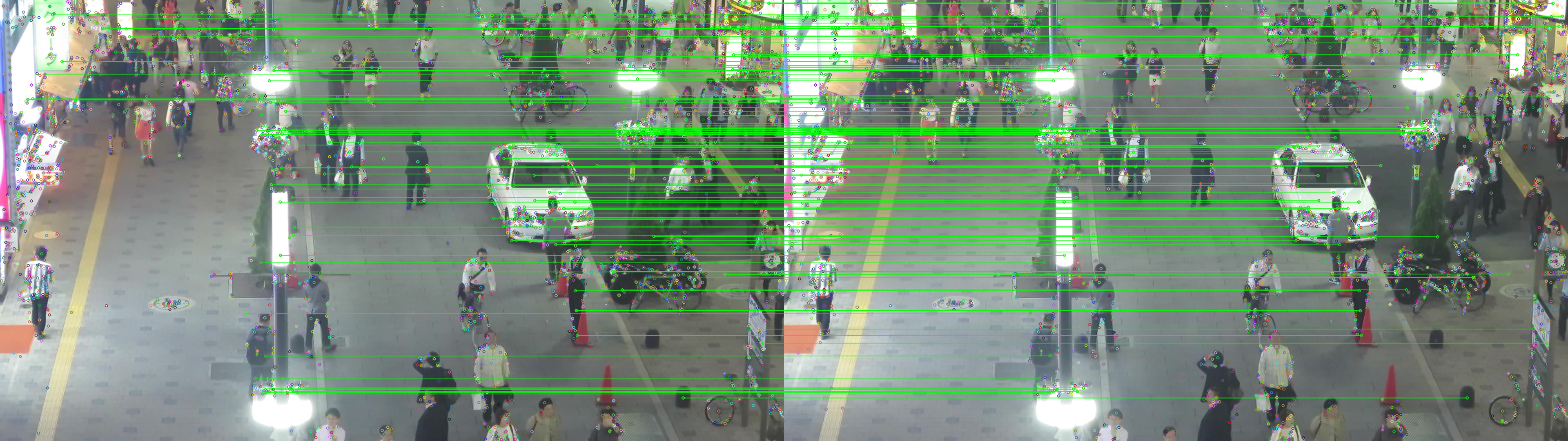}%
		\label{stationary}}
	\caption{The Visualization of Keypoints with Small Displacement as a Key Factor in the Proposed Camera Motion Detection. (a) The scarcity of keypoints when the camera is moving. (b) The abundance of keypoints when the camera is fixed.}
	\label{fig_motion}
\end{figure*}

In scenarios where the camera moves, the interpolation time-out should be decreased since linear interpolation requires the object trajectory to be linear, which is not true in such scenarios. So, this paper proposes a simple camera motion detection scheme. The camera is considered stationary when at least one keypoint shows a small displacement across two consecutive frames. Keypoints are extracted using ORB \cite{orb} and matched using a brute-force matcher, and the displacement of each keypoint is determined by the Euclidean distance of its coordinates across two frames. After examining five samples uniformly chosen from the video, a voting system determines whether the camera should be considered stationary throughout the entire video. Figure \ref{fig_motion} illustrates the importance of keypoints with small displacement in the proposed method. The green lines connect the same keypoint across two consecutive frames. In Figure \ref{moving}, the camera is moving, leaving only one keypoint with a displacement below \emph{15} pixels and no keypoints with a displacement below \emph{5} pixels. In contrast, Figure \ref{stationary} depicts a stationary camera, leading to an abundance of keypoints with a displacement below \emph{15} pixels and some keypoints with a displacement below \emph{5} pixels. Therefore, in this work, identifying a keypoint with a displacement below \emph{5} pixels indicates a stationary camera. 

\begin{figure*}[!t]
	\centering
	\subfloat[]{\includegraphics[width=1.75in]{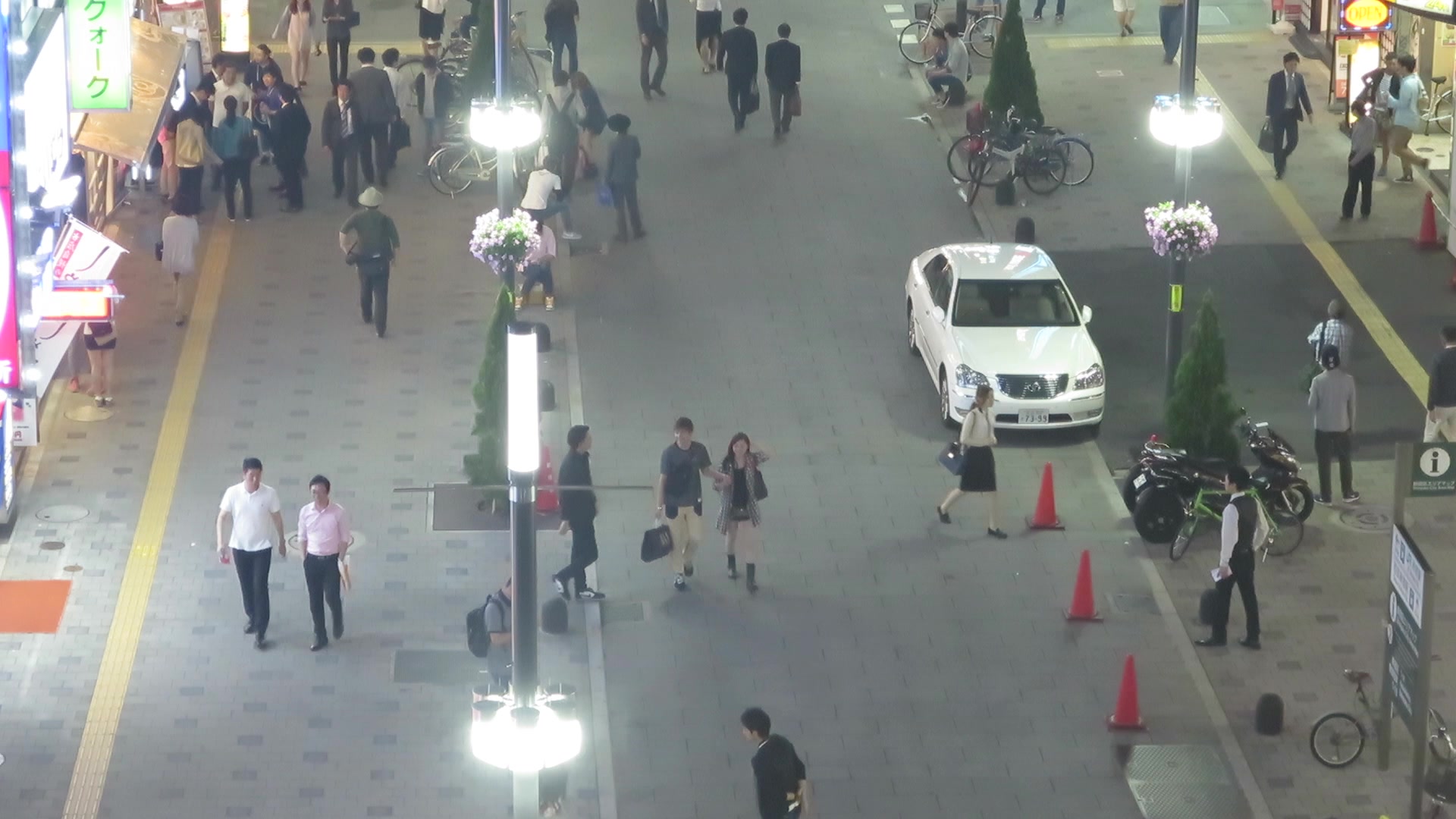}%
		\label{low}}
	\hfil
	\subfloat[]{\includegraphics[width=1.75in]{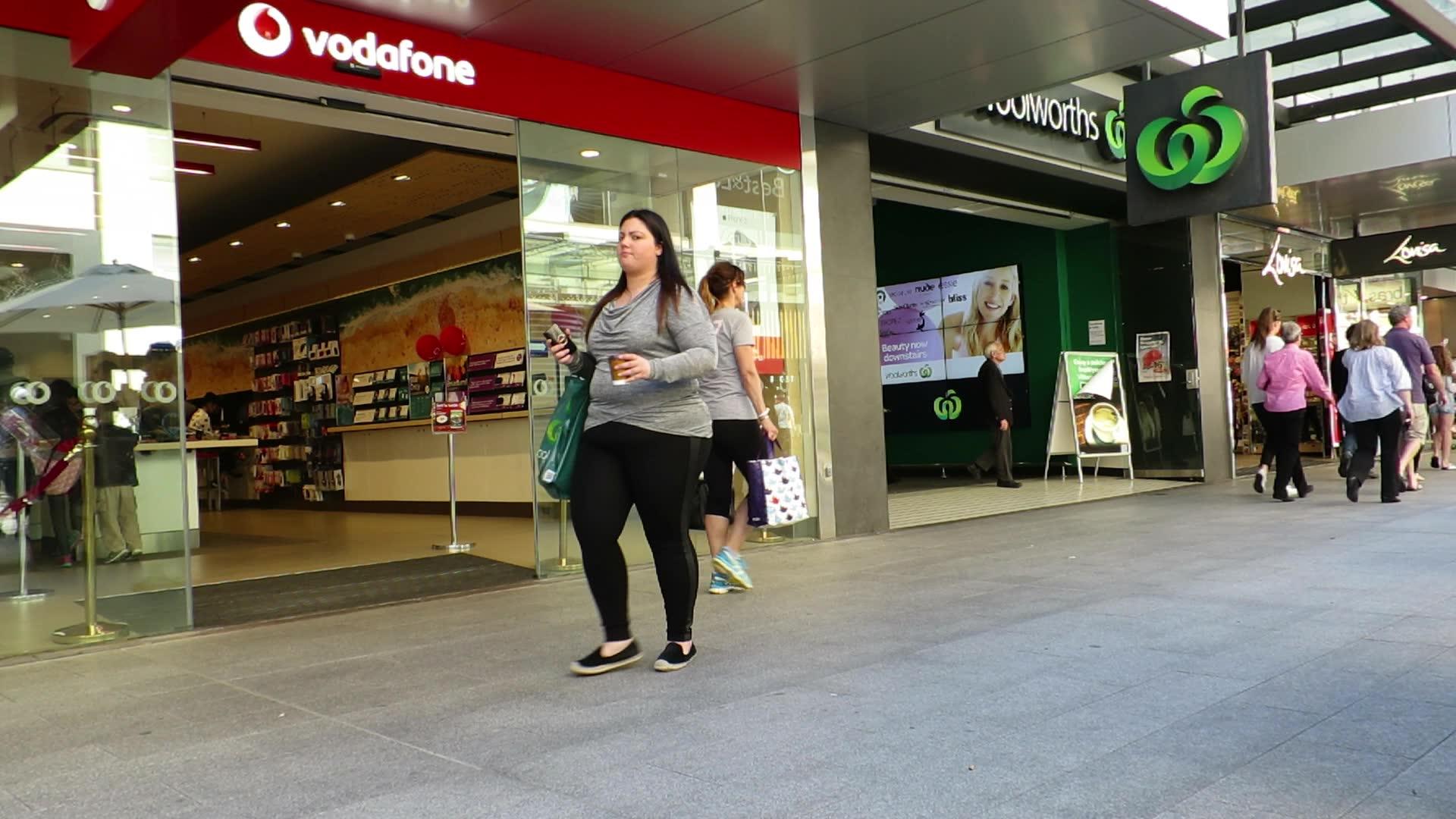}%
		\label{medium}}
	\hfil
	\subfloat[]{\includegraphics[width=1.75in]{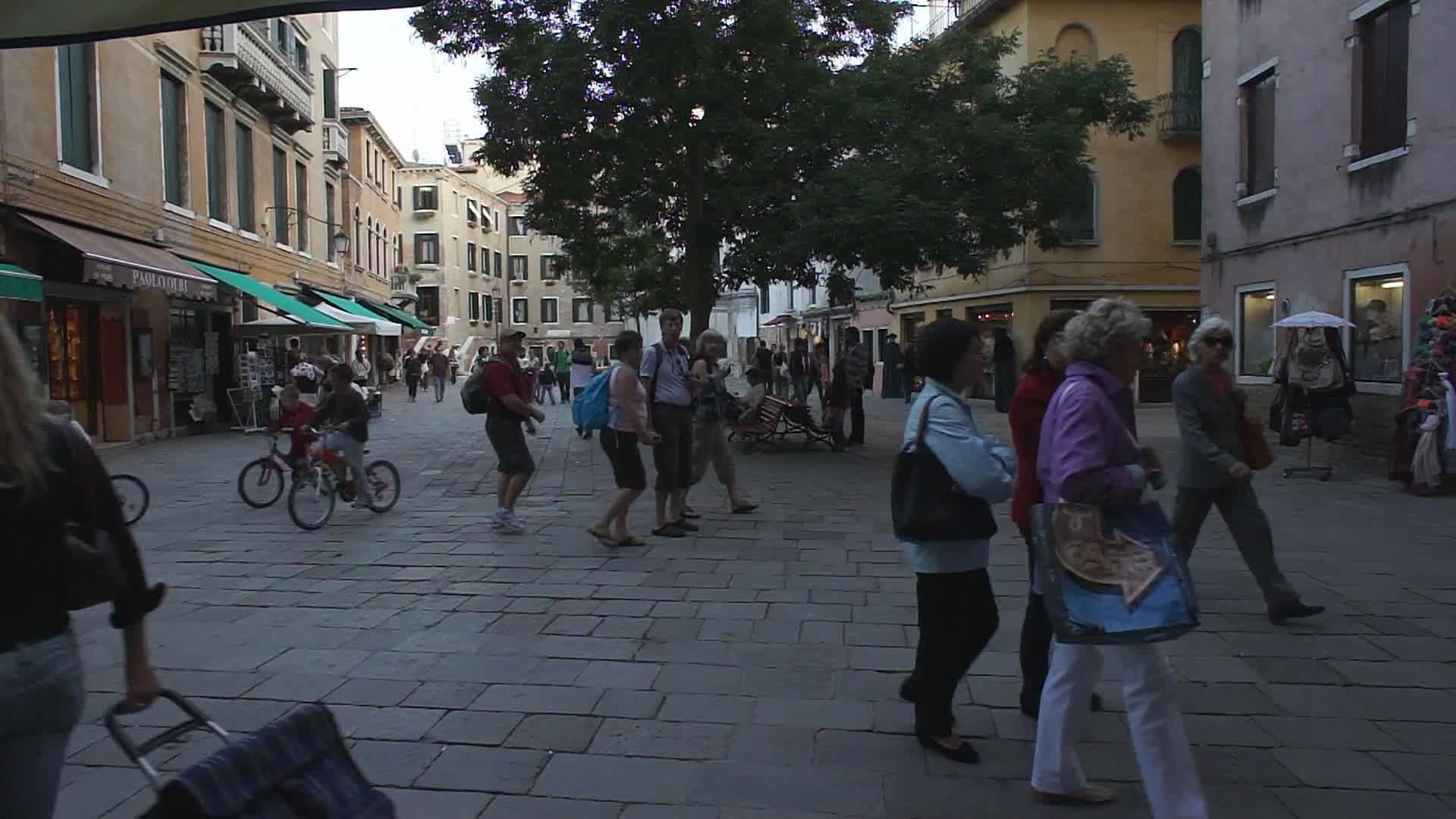}%
		\label{high}}
	\caption{Analyzing Scene Depth Using the Proposed Depth Score. (a) Depth score = 0.1. (b) Depth score = 0.56. (c) Depth score = 0.81}
	\label{fig_depth}
\end{figure*}

The depth of a scene impacts the interpolation time-out because objects located far from the camera are prone to be missed by the object detector, causing the linear trajectory assumption to fail. To discern deep scenes, this paper highlights the significant variation in the heights of same-class objects detected within such scenes. However, experiments demonstrate that the amount of height differences or the variance of heights is insufficient for effectively distinguishing a deep scene, as these measures are not resilient to noisy detections. Consequently, this paper proposes a depth estimation metric named the depth score, defined in Equation system \ref{eq-dmetric}.

\begin{subequations}\label{eq-dmetric}
	\begin{gather}
		h_{midrange} = \frac{h_{max} + h_{min}}{2}. \label{eq-dmetric-A} \\
		h_{average} = \frac{\sum_{i=1}^N h_{i}}{N}. \label{eq-dmetric-B}	\\
		depth\:score = \frac{|h_{average} - h_{midrange}|}{h_{midrange}}. \label{eq-dmetric-C}
	\end{gather}
\end{subequations}

In Equation \ref{eq-dmetric-A}, $h_{midrange}$ denotes the average height of the two objects with the largest and smallest heights. In Equation \ref{eq-dmetric-B}, $N$ represents the total number of detected objects, and $h_{average}$ stands for the average height across all detected objects. The proposed depth score always falls within the range $(0, 1)$: in a shallow scene, where the variation in object heights is minimal, $h_{midrange}$ approaches $h_{average}$ and results in a depth score of zero. Conversely, in a deep scene, the object detector often identifies large objects near the camera and misses many small objects farther away, causing $h_{average}$ to approach $h_{max}$. Consequently, the depth score approaches one in a deep scene, as depicted in Equation \ref{eq-dmetric-apprx}.

\begin{equation}
	\label{eq-dmetric-apprx}
		depth\:score \approx \frac{h_{max} - \frac{h_{max} + h_{min}}{2}}{\frac{h_{max} + h_{min}}{2}} = 1.
\end{equation}

In this study, a scene is classified as deep through the analysis of five samples uniformly chosen from the video. This classification relies on whether the average depth score for these samples surpasses a threshold determined using the validation set. Figure \ref{fig_depth} illustrates the depth scores for scenes with low, medium, and high depths, respectively. As shown in Figure \ref{fig_depth}, the proposed metric demonstrates effective performance.

Based on the discussed facts, Equation \ref{eq-nmin_final} represents the final formulation for determining the minimum length of a valid track. Similarly, Equation \ref{eq-ndti_final} represents the final formulation for determining the time-out duration for lost trajectory reconstruction by interpolation.

\begin{align}
	\label{eq-nmin_final}
	N_{min} =
	\begin{cases}
		C_{m1} \times FR,	& \text{Fixed = True, Deep = True}\\
		C_{m2} \times FR,	& \text{Fixed = True, Deep = False}\\
		C_{m3} \times FR,	& \text{Fixed = False, Deep = True}\\
		C_{m4} \times FR,	& \text{Fixed = False, Deep = False}	
	\end{cases}
\end{align}

\begin{align}
	\label{eq-ndti_final}
	N_{dti} =
	\begin{cases}
		C_{d1} \times FR,	& \text{Fixed = True, Deep = True}\\
		C_{d2} \times FR,	& \text{Fixed = True, Deep = False}\\
		C_{d3} \times FR,	& \text{Fixed = False, Deep = True}\\
		C_{d4} \times FR,	& \text{Fixed = False, Deep = False}	
	\end{cases}
\end{align}

In Equations \ref{eq-nmin_final} and \ref{eq-ndti_final}, $FR$ denotes the frame rate of the video. Furthermore, the term $Fixed$ shows a video captured from a stationary camera, while the term $Deep$ indicates a video portraying a deep scene. In Equation \ref{eq-nmin_final}, $N_{min}$ represents the minimum length required for valid tracks, while in Equation \ref{eq-ndti_final}, $N_{dti}$ denotes the interpolation time-out duration. The values of $C_{m1}$, $C_{m2}$, $C_{m3}$, $C_{m4}$, $C_{d1}$, $C_{d2}$, $C_{d3}$, and $C_{d4}$ are determined through experimentation on the validation set.

\section{Experiments}
\subsection{Experiment Setup}
\subsubsection{Datasets}
Experiments were conducted on the MOT17 \cite{mot17} and the MOT20 \cite{mot20} datasets, which cover diverse scenarios featuring varying perspectives, crowded environments, and camera motion states. This study specifically concentrates on the private setting of the datasets, allowing trackers to utilize their own object detectors.

The training data of MOT17 consists of $7$ videos totaling $5316$ frames, and its test data includes $7$ videos with a total of $5919$ frames. On average, each frame of the test set contains $31.8$ objects. The training data of MOT20 comprises 4 videos with a total of $8931$ frames, and its test data consists of $4$ videos totaling $4479$ frames. MOT20 focuses on crowded scenes, with an average of $170.9$ objects per frame in its test set.

MOT17 and MOT20 lack a dedicated validation set, leading previous studies to use the second half of the trainset videos for evaluation \cite{bytetrack, ocsort}. However, experimental findings show that events in one half of a video may not be as challenging as those in the other half. Therefore, this study adopts a 7-fold cross-validation approach, where one video is designated as the validation set and the remaining six serve as the training set in each iteration of the validation process.

\subsubsection{Metrics}
The evaluation metrics used in this work include HOTA \cite{hota}, IDF1 \cite{idf1}, DetA \cite{hota}, AssA \cite{hota}, and MOTA \cite{mota}. This work reports MOTA only to provide comprehensive results, but MOTA is considered an unfair metric due to its excessive bias toward detection quality. Currently, HOTA and IDF1 are recognized as metrics that provide a more equitable evaluation of trackers.

\subsubsection{Implementation Details}
The experiments were carried out utilizing the Python implementation of the proposed methods on a 2.20GHz Intel(R) Xeon(R) CPU. In instances requiring object detection, a Tesla T4 GPU was employed. Throughout the experiments, the input video frames were resized to dimensions of $800\times1440$. Evaluations were performed using the evaluation tool from \cite{trackeval}. Hyperparameter values, determined through experiments on the validation set, are summarized in Table \ref{tab_hyper_values}. The hyperparameters listed in the last row of Table \ref{tab_hyper_values} are not utilized in online tracking and are exclusive to offline tracking where post-processing is applied. As depicted in the table, across various tracking scenarios, $C_{m1}$, $C_{m2}$, $C_{m3}$, and $C_{m4}$ remain constant, indicating that the minimum valid track length originates from the behavior of the object detector. Therefore, these four hyperparameters can be merged into one.

\begin{table}
	\begin{center}
		\caption{Hyperparameter Values During Experiments}	
		\label{tab_hyper_values}
		\begin{tabular}{ccc}
			\toprule
			Hyperparameter & Value in MOT17 & Value in MOT20\\
			\midrule
			$LTH$ 		& 0.30	& 0.15 \\
			$MTH2$ 		& 0.10	& 0.30 \\					
			$HTH0$		& 0.82	& 0.70 \\
			$HTH_{m}$ 	& 0.10	& 0.07 \\
			$NTH0$		& 0.70	& 0.55 \\
			$NTH_{m}$ 	& 0.10	& 0.02 \\	
			$MTH0$ 		& 0.50	& 0.45 \\
			$MTH_{m}$ 	& 0.05	& 0.05 \\						
			$HMargin$ 	& 0.10	& 0.10 \\
			$VMargin$ 	& 0.10	& 0.15 \\
			$CTime$ 	& 1.00	& 1.00 \\
			$MTime$		& 0.70	& 0.50 \\
			\midrule
			$C_{m1}$, $C_{m2}$, $C_{m3}$, $C_{m4}$ 	& 1.0	& 1.5 \\		
			$C_{d1}$	& 0.7	& 0.5 \\
			$C_{d2}$	& 1.0	& 0.5 \\
			$C_{d3}$	& 0.1	& 0.5 \\
			$C_{d4}$	& 0.7	& 0.5 \\
			
			\bottomrule
		\end{tabular}
	\end{center}
\end{table}

\subsection{Ablation Studies}
This part examines the effects of each innovation introduced in this study through experiments. Due to restrictions limiting the number of experiments conducted on the test data of MOT17 to four, this study utilizes its predefined validation set of MOT17 for conducting ablation studies. Throughout these experiments, no post-processing is applied to the obtained results unless explicitly mentioned otherwise. The ablation study results are presented in Figure \ref{fig_cost} and Table \ref{tab_ablation}, and they are discussed in what follows.

\subsubsection{Cost Functions}
\begin{figure}[!t]
	\centering
	\includegraphics[width=\linewidth]{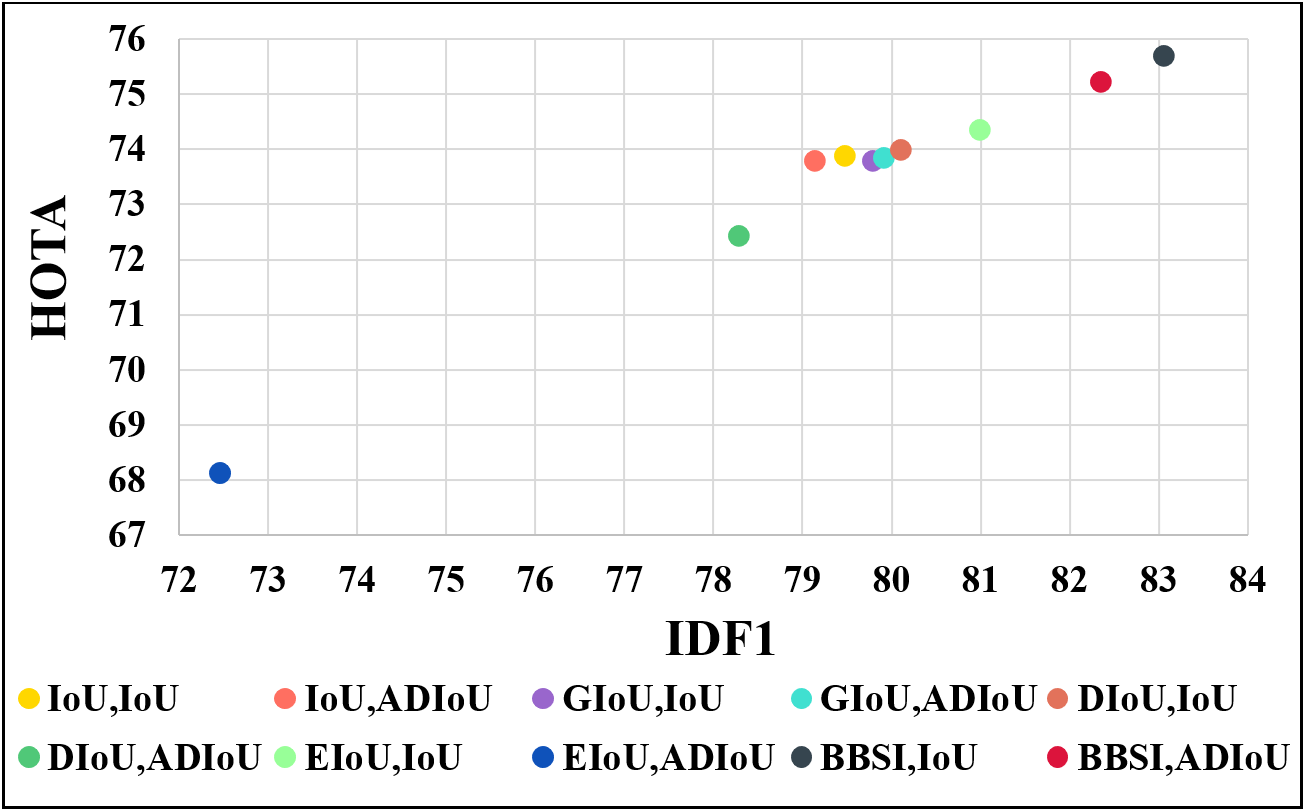}
	\caption{Results of Ablation Study on Various Association Cost Functions on the Defined Validation Set of MOT17.}
	\label{fig_cost}
\end{figure}

Figure \ref{fig_cost} presents the results of experiments utilizing various cost functions in association. In this figure, the legend below the chart implies the cost functions employed in the experiment. In the legend, the cost function on the left-hand side is utilized in the first association module, while the one on the right-hand side is employed in the second association module. Throughout all experiments, the tracker's hyperparameters are fine-tuned to optimize HOTA on the validation set. Moreover, no post-processing is applied to the results. Figure \ref{fig_cost} indicates the supremacy of the selected combination of association cost functions, BBSI and IoU, over its matches. 

\subsubsection{Time-out Based on Track Loss Location}
In Table \ref{tab_ablation}, the \emph{Default} mode represents the full-fledged implementation of the proposed tracker. The \emph{Same Time-out Everywhere} mode indicates an experiment in which the time-outs for both \emph{lost-at-center} and \emph{lost-at-margin} tracks are equal. As demonstrated in Table \ref{tab_ablation}, a tracker operating in the \emph{Default} mode outperforms a tracker in the \emph{Same Time-out Everywhere} mode in terms of HOTA and IDF1. The results of the \emph{Default} mode indicate that a lower time-out at margins helps reduce ID switching, thereby improving IDF1. Moreover, the higher time-out at the central area, as considered in the \emph{Default} mode, increases the possibility of revisiting lost tracks, consequently improving HOTA.

\subsubsection{Hyperparameter Tuning}
In Table \ref{tab_ablation}, the \emph{Fixed Hyperparameter} mode denotes an experiment where a tracker's hyperparameters are fixed at their optimal values and do not vary according to scene features. As indicated in Table \ref{tab_ablation}, a tracker in the \emph{Default} mode outperforms a tracker in the \emph{Fixed Hyperparameter} mode in terms of HOTA. Thus, assuming a dependency between hyperparameter values and scene features, introduced in Section \ref{features}, enhances tracking accuracy.

\subsubsection{Post-processing}
In Table \ref{tab_ablation}, the \emph{Simple Offline} mode represents an experiment where a basic linear interpolation is applied, while the \emph{Advanced Offline} mode corresponds to the same experiment with considerations for camera motion and scene depth. As evident in Table \ref{tab_ablation}, the introduction of post-processing leads to improvements across all metrics. However, in agreement with the expectations discussed in Section \ref{post}, a decrease in IDF1 in the \emph{Simple Offline} mode compared to the \emph{Advanced Offline} mode implies ID switching caused by the failure of the linear motion assumption. The results of the \emph{Advanced Offline} mode demonstrate that the innovations introduced in Section \ref{post} enhance the tracker's performance across all metrics.

\begin{table}
	\begin{center}
		\caption{Results of Ablation Study on Introduced Innovations on the Defined Validation Set of MOT17}
		\label{tab_ablation}	
		\begin{tabular}{cccc}
			\toprule
			Mode 	& HOTA & IDF1 & MOTA \\
			\midrule
			Default 					& 75.682 & 83.059 & 89.980 \\
			Same Time-out Everywhere	& 75.541 & 82.749 & 89.983 \\
			Fixed Hyperparameter 		& 75.606 & 83.193 & 90.027 \\	
			Simple Offline 				& 75.906 & 83.154 & 90.961 \\	
			Advanced Offline 			& 76.058 & 83.241 & 91.043 \\				
			\bottomrule
		\end{tabular}
	\end{center}
\end{table}

\subsection{Benchmark Results}

\begin{table*}
	\begin{center}
		\caption{Test Results on both MOT17 and MOT20 }
		\label{tab_results}	
		\begin{tabular}{ccccc|ccccc}
			\toprule
					& \multicolumn{4}{c}{MOT17}  & \multicolumn{4}{c}{MOT20} & \\
			Tracker	& HOTA & IDF1 & MOTA & Speed (Hz) & HOTA & IDF1 & MOTA & Speed (Hz) & Category\\
			\midrule
			This work (SFSORT)		 		& 61.7 & 74.4 & 78.8 & 2241.8	& 60.9 & 73.5 & 75.0 & 304.1&	\multirowcell{2}{real-time}\\
			TicrossNet \cite{TicrossNet}	& 57.1 & 69.1 & 74.7 & 32.6 	& 60.6 & 59.3 & 48.1 & 31.0	&	\\
			\midrule
			Semi-TCL \cite{semitcl}			& 59.8 & 73.2 & 73.3 & 88.8 	& 55.3 & 70.1 & 65.2 & 22.4 & \multirowcell{2}{near real-time}\\		
			RetinaMOT \cite{RetinaMOT}		& 58.1 & 70.9 & 74.1 & 67.5 	& 54.1 & 67.5 & 66.8 & 22.4	&	\\
			\midrule		
			SGTMOT \cite{SGTMOT}			& 60.6 & 72.4 & 76.3 & 62.5 	& 56.9 & 70.5 & 72.8 & 17.2 & 
			\multirowcell{11}{non-real-time}\\				
			BASE \cite{BASE}				& 64.5 & 78.6 & 81.9 & 331.3	& 63.5 & 77.6 &	78.2 & 16.8	&	\\
			TransTrack	\cite{TransTrack}	& 54.1 & 63.5 & 75.2 & 59.2 	& 48.9 & 59.4 & 65.0 & 14.9 &	\\
			MAA	\cite{maatrack}				& 62.0 & 75.9 & 79.4 & 189.1	& 57.3 & 71.2 & 73.9 & 14.7	&	\\
			BYTEv2 \cite{byte2}				& 63.6 & 78.9 & 80.6 & 48.2 	& 61.4 & 75.6 & 77.3 & 11.9 &	\\	
			CrowdTrack \cite{CrowdTrack}	& 60.3 & 73.6 & 75.6 & 140.8 	& 55.0 & 68.2 & 70.7 & 9.5 	&	\\
			FineTrack \cite{FineTrack}		& 64.3 & 79.5 & 80.0 & 35.5 	& 63.6 & 79.0 & 77.9 & 9.0	&	\\			
			QDTrack	\cite{QDTrack}			& 63.5 & 77.5 & 78.7 & 37.0 	& 60.0 & 73.8 & 74.7 & 7.5 	&	\\
			ImprAsso \cite{imprasso}		& 66.4 & 82.1 & 82.2 & 143.7	& 64.6 & 78.6 & 78.8 & 6.4	&	\\		
			StrongTBD \cite{strongtbd}		& 65.6 & 80.8 & 81.6 & 111.9 	& 63.6 & 77.0 & 78.0 & 2.9	&	\\
			SelfAT \cite{SelfAT}			& 64.4 & 79.8 & 80.0 & 142.1	& 62.6 & 76.6 & 75.0 & 1.2	&	\\			
			\bottomrule	
		\end{tabular}
	\end{center}
\end{table*}

Table \ref{tab_results} presents a comprehensive comparison of the proposed tracker alongside a selection of its counterparts, focusing on key accuracy metrics such as MOTA, IDF1, and HOTA. The comparison includes trackers capable of real-time processing of videos from the MOT17 dataset. As evident in Table \ref{tab_results}, for both MOT17 and MOT20 datasets, besides the proposed tracker, only one other tracker reported a tracking speed exceeding 30 Hz, the minimum threshold for real-time tracking. SFSORT, as shown in Table \ref{tab_results}, exceeds the processing speed of the fastest previous tracker on crowded videos of the MOT20 dataset by nearly tenfold and is almost seven times faster than the leading previous tracker on MOT17 dataset videos. Furthermore, Table \ref{tab_results} indicates that SFSORT exhibits superior accuracy across all evaluated metrics, including HOTA, IDF1, and MOTA, compared to any other real-time or near real-time tracker.
 
\subsection{Results of Similar Trackers under Identical Exprimental Conditions}
Given the influence of both hardware and software on the results, a fair comparison is ensured by running the official Python implementation of each tracker on identical hardware. Performance optimization is attained by adjusting the hyperparameters of each tracker using a genetic algorithm-based approach proposed in \cite{boxmot}. This part compares the accuracy and speed of the proposed multi-object tracking system with a selection of its counterparts. The selection comprises the top four SORT-based trackers in terms of HOTA.

To demonstrate the adaptability of the proposed tracking system to various object detectors, a different object detector is used in the experiments in this part compared to the one used previously. The study by \cite{yolo7} introduces YOLOv7 as a high-speed, accurate object detector that outperforms prior detectors such as \cite{fasterrcnn, cascadercnn, iounet, yolo, centernet, fcos, yolox}, commonly used in trackers. Given that YOLOv8, a new object detector presented in \cite{yolo8}, surpasses YOLOv7 in terms of accuracy, this part employs YOLOv8 as the object detector in the tracking system. The experiment will utilize YOLOv8n, the most computationally lightweight version of the YOLOv8 detector family. This version is capable of running on a wide range of processors. 

\begin{figure}[!t]
	\centering
	\includegraphics[width=\linewidth]{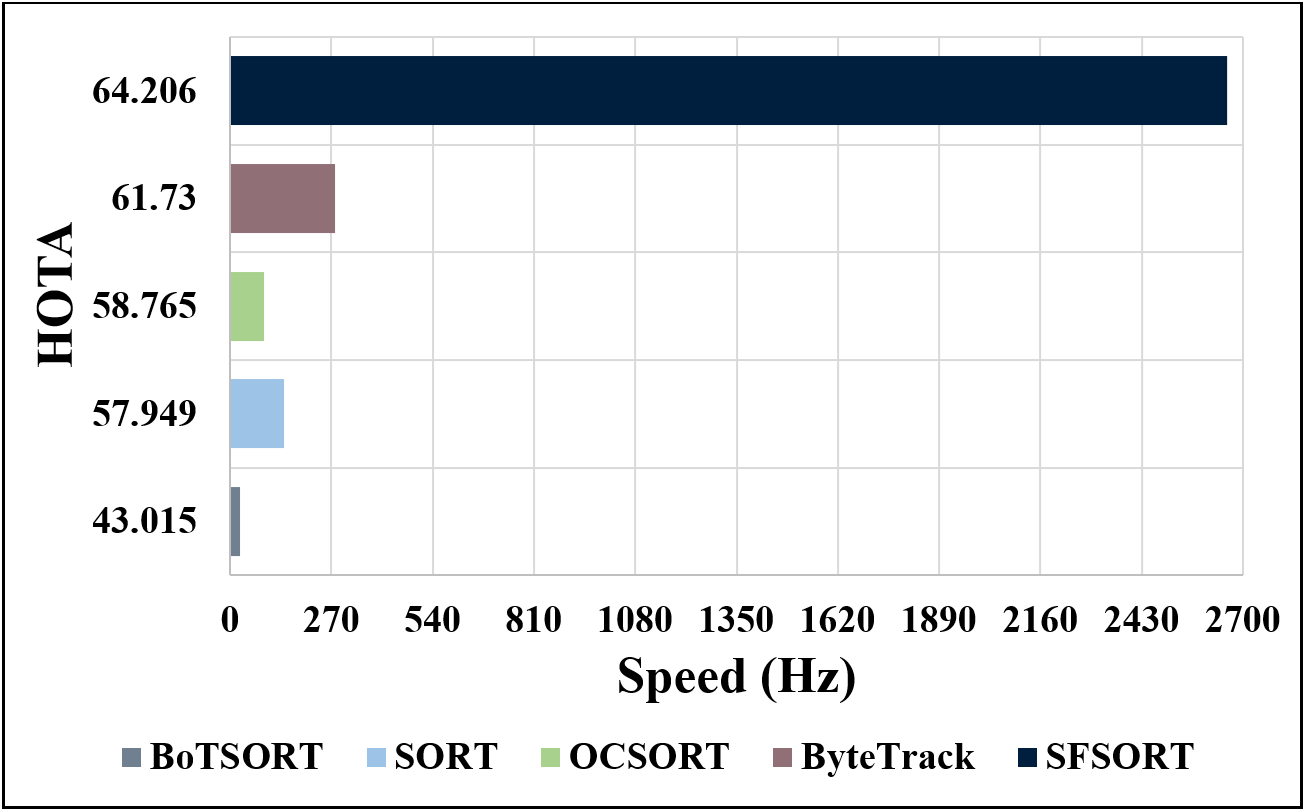}
	\caption{Comparison of Accuracy and Speed for SORT-Based Trackers under Identical Experimental Conditions.}
	\label{fig_hotafps}
\end{figure}

Figure \ref{fig_hotafps} presents the evaluation results, affirming that SFSORT outperforms its competitors in terms of both accuracy and speed. The reported speed in Figure \ref{fig_hotafps} specifically refers to the tracking algorithm's performance, excluding the object detector's processing time. The source code for SFSORT, the fine-tuned YOLOv8 object detection model, as well as examples and tutorials, can be accessed at \url{https://github.com/gitmehrdad/SFSORT}.

\section{Conclusion}
This paper introduces SFSORT, a real-time multi-object tracking system based on scene features. The system's computational efficiency makes it suitable for various platforms, ranging from edge processors to cloud servers. While most previous works have focused on enhancing accuracy, it is crucial to recognize the importance of speed, especially in practical applications where deploying expensive hardware may not be feasible. Notably, SFSORT relies solely on scene features and object bounding box properties, eliminating the need for the original frame and making it suitable for privacy-preserving applications. The Python implementation of SFSORT achieves an impressive operational speed of $2242$ Hz, making it the world's fastest multi-object tracker and one of two trackers capable of real-time operation on both normal and crowded scenes. In evaluations using the MOT17 dataset, SFSORT achieved an HOTA of $61.7$\%, with further improvements anticipated through post-processing. Future enhancements may involve the integration of trajectory predictors like the Kalman Filter, the utilization of deep features such as ReID neural networks, and the implementation of Camera Motion Compensation to enhance tracking accuracy.

\end{document}